# Decentralized Control of Cooperative Systems: Categorization and Complexity Analysis

**Claudia V. Goldman**                                        CLAG@CS.UMASS.EDU
**Shlomo Zilberstein**                                        SHLOMO@CS.UMASS.EDU
*Department of Computer Science,*
*University of Massachusetts Amherst, Amherst, MA 01003 USA*

## Abstract

Decentralized control of cooperative systems captures the operation of a group of decision-makers that share a single global objective. The difficulty in solving optimally such problems arises when the agents lack full observability of the global state of the system when they operate. The general problem has been shown to be NEXP-complete. In this paper, we identify classes of decentralized control problems whose complexity ranges between NEXP and P. In particular, we study problems characterized by independent transitions, independent observations, and goal-oriented objective functions. Two algorithms are shown to solve optimally useful classes of goal-oriented decentralized processes in polynomial time. This paper also studies information sharing among the decision-makers, which can improve their performance. We distinguish between three ways in which agents can exchange information: indirect communication, direct communication and sharing state features that are not controlled by the agents. Our analysis shows that for every class of problems we consider, introducing direct or indirect communication does not change the worst-case complexity. The results provide a better understanding of the complexity of decentralized control problems that arise in practice and facilitate the development of planning algorithms for these problems.

## 1. Introduction

Markov decision processes have been widely studied as a mathematical framework for sequential decision-making in stochastic domains. In particular, single-agent planning problems in stochastic domains were modeled as partially observable Markov decision processes (POMDPs) or fully-observable MDPs (Dean, Kaelbling, Kirman, & Nicholson, 1995; Kaelbling, Littman, & Cassandra, 1998; Boutilier, Dearden, & Goldszmidt, 1995). Borrowing from Operations Research techniques, optimal plans can be computed for these planning problems by solving the corresponding Markov decision problem. There has been a vast amount of progress in solving individual MDPs by exploiting domain structure (e.g., Boutilier et al., 1995; Feng & Hansen, 2002). Approximations of MDPs have also been studied, for example, by Guestrin et al. (2003), assuming that the reward function can be decomposed into local reward functions each depending on only a small set of variables.

We are interested in a single Markov decision process that is collaboratively controlled by multiple decision-makers. The group of agents cooperate in the sense that they all want to maximize a single global objective (or minimize the cost of achieving it). Nevertheless, the decision-makers do not have full observability of the whole system at the time of execution. These processes can be found in many application domains such as multi-robot problems, flexible manufacturing and information gathering. For example, consider a group of space exploration rovers, like those sent by NASA to Mars.[1] These rovers could be assigned a set of experiments to perform on the planet before they need to meet. They may have a limited amount of time to perform these experiments. Then, the robots need to decide what experiments to perform and how much time they should invest in each one given the available battery power and remaining time. Decentralized cooperative problems also include information gathering systems, where several software agents with access to different servers may provide answers to a user's query. These agents' global objective is to give the user the best answer as early as possible, given the load on their servers and the preferences given by the user. Controlling the operation of the agents in such examples is not trivial because they face uncertainty regarding the state of the environment (e.g., the load on the communication link between the servers in the information gathering example may vary), the outcome of their actions (e.g., the rovers on Mars may be uncertain about their location and time needed to move from one target location to another), and the accuracy of their observations. All these types of uncertainty are taken into account when solving such decentralized problems.

The processes described above are examples of decentralized partially-observable Markov decision processes (Dec-POMDPs) or decentralized Markov decision processes (Dec-MDPs).[2] The complexity of solving these problems has been studied recently (Bernstein, Givan, Immerman, & Zilberstein, 2002; Pynadath & Tambe, 2002). Bernstein et al. have shown that solving optimally a Dec-MDP is NEXP-complete by reducing the control problem to the tiling problem. Rabinovich et al. (2003) have shown that even approximating the off-line optimal solution of a Dec-MDP remains NEXP. Nair et al. (2003) have presented the Joint Equilibrium-based Search for Policies (JESP) algorithm that finds a *locally-optimal* joint solution. Researchers have attempted to *approximate* the coordination problem by proposing *on-line* learning procedures. Peshkin et al. (2000) have studied how to approximate the decentralized solution based on a gradient descent approach for on-line learning (when the agents do not know the model). Schneider et al. (1999) assume that each decision-maker is assigned a local optimization problem. Their analysis shows how to approximate the global optimal value function when agents may exchange information about their local values at no cost. Neither convergence nor bounds have been established for this approach. Wolpert et al. (1999) assume that each agent runs a predetermined reinforcement learning algorithm and transforms the coordination problem into updating the local reward function so as to maximize the global reward function. Again, this is an approximation algorithm for on-line learning that does not guarantee convergence. Agents in this model may communicate freely. Guestrin et al. study off-line approximations (a centralized approach (Guestrin, Koller, & Parr, 2001) and a distributed approach (Guestrin & Gordon, 2002)), where a known structure of the agents' action dependencies induces a message passing structure. In this context, agents choose their actions in turns and communication is free. The solution is based on the assumption that the value function of the system can be represented by a set of compact basis functions, which are then approximated. The complexity of the algorithm is exponential in the width of the coordination graph. The order of elimination is needed beforehand because it has a great effect on the result.

---

1. mars.jpl.nasa.gov/mer/
2. These problems are defined in Definition 4 in Section 2.

What is common to these approaches is the departure from the assumption that each agent has a known local reward function. The questions that they attempt to answer, hence, take the form of how to design or manipulate local reward functions so as to approximate the actual system reward function.

In this work, we take a different approach. We aim at solving the decentralized control problem off-line without making any particular assumptions about the reward functions of each agent. The problem is therefore analyzed from a decentralized perspective. We have developed a formal model for decentralized control, which extends current models based on Markov decision processes. We refer to the most general problem where information sharing between the agents can result from indirect communication (i.e., via observations), by direct communication (i.e., via messages) or by sharing common uncontrollable features of the environment (defined in Section 2.2). When direct communication is possible, we assume that communication may incur a cost. Communication can assist the agents to better control the process, but it may not be possible or desirable at every moment. Exchanging information may incur a cost associated with the required bandwidth, the risk of revealing information to competing agents or the complexity of solving an additional problem related to the communication (e.g., computing the messages). Assuming that communication may not be reliable adds another dimension of complexity to the problem.

Becker et al. (2003) presented the first algorithm for optimal off-line decentralized control when a certain structure of the joint reward was assumed. Recently, Hansen et al. (2004) showed how to generalize dynamic programming to solve optimally general decentralized problems. Nevertheless, no existing technique solves efficiently special classes of Dec-POMDPs that we identify in this paper. Pynadath and Tambe (2002) studied a similar model to ours, although they did not propose an algorithm for solving the decentralized control problem. Claus and Boutilier (1998) studied a simple case of decentralized control where agents share information about each other's actions during the off-line planning stage. The solution presented in their example includes a joint policy of a single action for each agent to be followed in a stateless environment. The agents learn which equilibrium to play. In our model, partial observability is assumed and the scenarios studied are more complex and include multiple states. Centralized multi-agent systems (MAS) were also studied in the framework of MDPs (e.g., Boutilier, 1999), where both the off-line planning stage and the on-line stage are controlled by a central entity, or by all the agents in the system, who have full observability.

Coordination and cooperation have been studied extensively by the distributed artificial intelligence community (Durfee, 1988; Grosz & Kraus, 1996; Smith, 1988) assuming a known and fixed language of communication. KQML (Finin, Labrou, & Mayfield, 1997) is an example of one standard designed to specify the possible communication between the agents. Balch and Arkin's (1994) approach to communication between robots is inspired by biological models and refers to specific tasks such as foraging, consumption and grazing. Their empirical study was performed in the context of reactive systems and communication was free. Our aim is to find optimal policies of communication and action off-line, taking into account information that agents can acquire on-line. Game theory researchers (Luce & Raiffa, 1957; Aumann & Hart, 1994) have also looked at communication, although the approaches and questions are somewhat different from ours. For example, Wärneryd (1993), and Blume and Sobel (1995) study how the receiver of a message may alter its actions in games where only one agent can send a single message at no cost. In contrast, we study sequential decision-making problems where all the agents can send various types of messages, which incur some cost. Our objective is to analyze the complexity of the problem and formulate algorithms that can find optimal policies of behavior as well as communication.

Our work focuses on decentralized cooperative MAS. Agents in cooperative MAS typically have limited ability to share information during execution (due to the distributed nature of the system). However, due to the cooperative nature of such systems, these constraints rarely apply to the pre-execution stage. Thus, we focus on cooperative agents that can share information fully at the off-line planning stage. Unlike the centralized approach, these agents will be acting in real-time in a decentralized manner. They must take this into account while planning off-line.

Sub-classes of Dec-POMDPs can be characterized based on how the global states, transition function, observation function, and reward function relate to the partial view of each of the controlling agents. In the simplest case, the global states can be factored, the probability of transitions and observations are independent, the observations combined determine the global state, and the reward function can be easily defined as the sum of local reward functions. In this extreme case we can say that the Dec-POMDP is equivalent to the combination of $n$ independent MDPs. This simple case is solvable by combining all the optimal solutions of the independent MDPs. We are interested in more complex Dec-POMDPs, in which some or all of these assumptions are violated. In particular, we characterize Dec-POMDPs, which may be jointly fully-observable, may have independent transitions and observations and may result in goal-oriented behavior. We analyze the complexity of solving these classes off-line and show that it ranges from NEXP to P. We also identify different forms of information sharing and show that when direct communication is possible, exchanging local observations is sufficient to attain optimal decentralized control.

The contributions of the paper are as follows: formalizing special classes of decentralized control problems (Section 2), identifying classes of decentralized control that are critical in decreasing the complexity of the problem (Section 3), designing algorithms for controlling optimally a decentralized process with goal-oriented behavior (Section 4), and extending the formal framework by introducing direct communication, considering the trade-off between its cost and the value of the information acquired, and analyzing the complexity of solving optimally such problems (Section 5).

## 2. The Dec-POMDP Model

We are interested in a stochastic process that is cooperatively controlled by a group of decision-makers who lack a central view of the global state. Nevertheless, these agents share a set of objectives and all of them are interested in maximizing the utility of the system. The process is decentralized because none of the agents can control the whole process and none of the agents has a full view of the global state. The formal framework in which we study such decentrally controlled processes, called Dec-POMDPs, is presented below (originally presented by Bernstein et al., 2002). For simplicity of exposition, the formal model is presented for two agents, although it can be extended to any number.

$M = \langle S, A_1, A_2, P, R, \Omega_1, \Omega_2, O, T \rangle$ where:

- $S$ is a finite set of world states with a distinguished initial state $s^0$.

- $A_1$ and $A_2$ are finite sets of control actions. $a_i$ denotes an action performed by agent $i$.

- $P$ is the transition probability function. $P(s'|s, a_1, a_2)$ is the probability of moving from state $s \in S$ to state $s' \in S$ when agents 1 and 2 perform actions $a_1$ and $a_2$ respectively. We note that the transition model is stationary, i.e., it is independent of time.

- $R$ is the global reward function. $R(s, a_1, a_2, s')$ represents the reward obtained by the system as a whole, when agent 1 executes action $a_1$ and agent 2 executes action $a_2$ in state $s$ resulting in a transition to state $s'$.

- $\Omega_1$ and $\Omega_2$ are finite sets of observations.

- $O$ is the observation function. $O(o_1, o_2|s, a_1, a_2, s')$ is the probability of observing $o_1$ and $o_2$ (respectively by the two agents) when in state $s$ agent 1 takes action $a_1$ and agent 2 takes action $a_2$, resulting is state $s'$.

- If the Dec-POMDP has a finite horizon, it is represented by a positive integer $T$.

We will illustrate our definitions and results through the Meeting under Uncertainty example. In this scenario, we assume for simplicity that there are two robots operating on a two-dimensional grid. The state of the system is given by the locations of each one of the robots, $s = [(x_1, y_1)(x_2, y_2)]$. In a more general example, this state may include other features such as the topology of the terrain, which no rover may be able to sense (e.g., due to lack of equipment). The robots cannot observe each other, and the movement actions they can perform have uncertain outcomes (e.g., each robot will successfully move to the next location with some probability, but it may remain at the same location where it took the action). In a more uncertain case, the topology of the terrain, although hidden from the agents' observations may affect the resulting location of a moving action. The robots' objective is to minimize the time to meet. The observation of robot $i$ corresponds to $i$'s $x$ and $y$ coordinates. Solving optimally such a decentralized problem means finding the sequence of moves for each agent such that they meet as soon as possible.

Given the Dec-POMDP model, a *local policy* of action for a single agent is given by a mapping from sequences of observations to actions. In our example, a robot's local policy instructs it to take a certain movement action given the sequence of locations it has observed so far. A *joint policy* is a tuple composed of these local policies, one for each agent. To solve a decentralized POMDP problem one must find the optimal joint policy that is, the one with maximum value (for example given by the maximum expected accumulated global reward). By associating a small negative reward with each action, maximizing reward coincides with the objective of minimizing time to meet. Notice that the agents' observations can be dependent on each other, allowing the agents to *know* what other agents are observing and in some sense enabling the agents to obtain full observability of the system state. That is, even though the agents may not communicate directly, when the observations are dependent, agents may be able to obtain information about the others without receiving direct messages. For example, assume that in our scenario there are certain locations, which can host only one robot at a time. If one robot observes that it is located at any one of these sites, then it knows that the other robot cannot be located there even though this robot does not actually see the other nor receive any information from it.

In the next section, we characterize certain properties that a decentralized process may have. These properties will play an important role when analyzing the complexity of solving different classes of decentrally controlled cooperative problems.

## 2.1 Classes of Dec-POMDPs

It is known that solving optimally *general* decentralized problems is very hard. We are interested in identifying interesting subclasses of the general problem and their characteristics. As we show in Section 3, this classification reveals interesting complexity results and facilitates the design of algorithms. The first two categories that we define involve independence of the transitions or the observations of the agents. Figure 1 presents a snapshot of the system at times $t$ and $t + 1$ that illustrates these categories. We assume that the global states are factored and we denote by $s_i$ the features of the world that may be observed fully by agent $i$.[3] Specifically, agent $i$ senses observation $o_i$, which in some cases can be identical to the local state of the agent. As a result of agent $i$ performing action $a_i$, when the system is at local state $s_i$, the system transitions to state $s'_i$, where the agent observes $o_i$. The resulting state $s'_i$ is not affected by the resulting state of the other agent $s'_j$. Similarly, the observations in these resulting states do not influence each other given the corresponding local information.

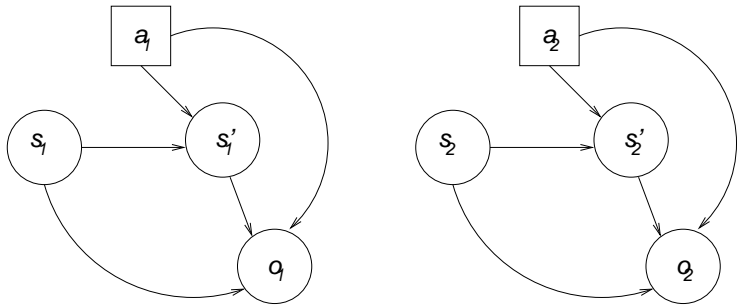

Figure 1: Independent Transitions and Observations.

The formal definitions for decentralized processes with independent transitions and observations follow.

**Definition 1 (Dec-POMDP with Independent Transitions)** *A Dec-POMDP has independent transitions if the set $S$ of states can be factored into two components $S_1$ and $S_2$ such that:*

$$\forall s_1, s'_1 \in S_1, \forall s_2, s'_2 \in S_2, \forall a_1 \in A_1, \forall a_2 \in A_2,$$

$$Pr(s'_1|(s_1, s_2), a_1, a_2, s'_2) = Pr(s'_1|s_1, a_1) \ \wedge$$

$$Pr(s'_2|(s_1, s_2), a_1, a_2, s'_1) = Pr(s'_2|s_2, a_2).$$

---

3. A partial view in general does not have to correspond to a partition of the global state, but we limit ourselves to problems with factored representations.

*In other words, the transition probability $P$ of the Dec-POMDP can be represented as $P = P_1 \times P_2$, where $P_1 = Pr(s_1'|s_1, a_1)$ and $P_2 = Pr(s_2'|s_2, a_2)$.*

The Meeting under Uncertainty example can be modified to present a problem with dependent transitions. For example, the scenario may include obstacles that can be moved by the agents when performing a *move-obstacle* action. The outcome of this action is that the rover's location is updated based on the direction of the move. For example, assume that one rover is located at (1,1) and finds an obstacle in location (1,2) which blocks its movement towards some goal. The rover decides to perform a move-obstacle action on that obstacle to the east. This action may have a stochastic outcome, for example with some probability $P_{obstacle}$ the rover succeeds and with probability $1 - P_{obstacle}$ it remains in the same location (1,1). With this modification, the problem has dependent transitions. For example, if both rovers decide to apply a move-obstacle action to the same obstacle, standing each on different sides of the obstacle, then the resulting state of each rover depends on the other rover also performing a move-obstacle action.

Moreover, the observations of the agents can be independent, i.e., each agent's own observations are independent of the other agents' actions.

**Definition 2 (Dec-POMDP with Independent Observations)** *A Dec-POMDP has independent observations if the set $S$ of states can be factored into two components $S_1$ and $S_2$ such that:*

$$\forall o_1 \in \Omega_1, \forall o_2 \in \Omega_2, \forall s = (s_1, s_2), s' = (s_1', s_2') \in S, \forall a_1 \in A_1, \forall a_2 \in A_2,$$

$$Pr(o_1|(s_1, s_2), a_1, a_2, (s_1', s_2'), o_2) = Pr(o_1|s_1, a_1, s_1') \wedge$$

$$Pr(o_2|(s_1, s_2), a_1, a_2, (s_1', s_2'), o_1) = Pr(o_2|s_2, a_2, s_2')$$

$$O(o_1, o_2|(s_1, s_2), a_1, a_2, (s_1', s_2')) =$$

$$Pr(o_1|(s_1, s_2), a_1, a_2, (s_1', s_2'), o_2) \times Pr(o_2|(s_1, s_2), a_1, a_2, (s_1', s_2'), o_1).$$

*In other words, the observation probability $O$ of the Dec-POMDP can be decomposed into two observation probabilities $O_1$ and $O_2$, such that $O_1 = Pr(o_1|(s_1, s_2), a_1, a_2, (s_1', s_2'), o_2)$ and $O_2 = Pr(o_2|(s_1, s_2), a_1, a_2, (s_1', s_2'), o_1)$.*

In the Meeting under Uncertainty example, if a robot's observation of its current location depends only on its transition from its previous location and on the action it performed, then the observations are independent. But more complex problems can arise if each agent's observation depends also on the other agent's location or action. For example, assume that in addition to the move actions, the rovers can also turn a flashlight that they hold on or off. In such cases, one rover's observation may be affected by the light or the lack of light in the area where the rover is observing. That is, the probability of a certain observation $o$ depends on both rovers' actions and states.

Throughout the paper, when we refer to a Dec-POMDP with independent transitions and observations, we assume the same decomposition of the global states into $S_1$ and $S_2$. We refer to $S_i$ as the *partial view* of agent $i$.

There are cases where agents may observe some common features of the global state, leading to dependent observations. When these common features have no impact on the transition model or reward, the problem can be reformulated to satisfy the property of independent observations. Such reformulation is shown to reduce the complexity of the problem in Section 3.

One of the main difficulties in solving Dec-POMDPs optimally results from the lack of full observability of the complete global state. An agent has full observability if it can determine with certainty the global state of the world from its local observation. For example, each time a robot observes where it is located, it also observes the other robot's location. Knowing both locations enables both agents to make the optimal decision about their next move in order to eventually meet sooner.

**Definition 3 (Fully-observable Dec-POMDP)** *A Dec-POMDP is* fully observable *if there exists a mapping for each agent $i$, $F_i : \Omega_i \rightarrow S$ such that whenever $O(o_1, o_2 | s, a_1, a_2, s')$ is non-zero then $F_i(o_i) = s'$.*

This paper analyzes decentralized problems where full observability is not satisfied. Instead, we distinguish between two classes of problems with restricted system observability: (1) *combining* both agents' partial views leads to the complete global state, and (2) each agent's *own* partial view $s_i$ is fully observable.[4] We say that Dec-POMDPs with property (1) are *jointly fully observable.*

**Definition 4 (Jointly Fully-observable Dec-POMDP)** *A Dec-POMDP is* jointly fully observable *(also referred to as a Dec-MDP) if there exists a mapping $J : \Omega_1 \times \Omega_2 \rightarrow S$ such that whenever $O(o_1, o_2 | s, a_1, a_2, s')$ is non-zero then $J(o_1, o_2) = s'$.*

Notice that both Definitions 1 and 2 apply to Dec-MDPs as well as to Dec-POMDPs. The Meeting under Uncertainty scenario presented in Section 2 when the state includes only the rovers' locations (and it does not include the terrain topology) is actually a Dec-MDP. The global state is given by the two pairs of coordinates. There is no other feature in the system state that is hidden from the agents. Notice that even though the combination of the agents' observations results in the global state, each agent may still be uncertain about its own current partial view. Each agent may have a belief about its actual location. We define another class of problems, *locally fully-observable* Dec-POMDPs, where each agent is certain about its observations. General Dec-MDPs consider only the *combination* of the agents' observations, but the definition does not say anything about *each* agent's observation.

**Definition 5 (Locally Fully-observable Dec-POMDP)** *A Dec-POMDP with independent transitions is* locally fully observable *if there exists a mapping for each agent $i$, $L_i : \Omega_i \rightarrow S_i$ such that whenever $O(o_1, o_2 | s, a_1, a_2, s')$ (where $s = (s_1, s_2)$ and $s' = (s'_1, s'_2)$) is non-zero then $L_1(o_1) = s'_1$ and $L_2(o_2) = s'_2$, where $s_i, s'_i \in S_i$ are the partial views of agent $i$.*

The Meeting under Uncertainty example is locally fully observable because each robot knows with certainty where it is located. We may think of more realistic robots, which may be uncertain about their actual location due to hardware inaccuracies.

---

4. This case does not induce full observability as defined in Definition 3 since there may be information in the global state that is hidden from the agents.

Notice that a jointly fully-observable process, which is also locally fully observable is not necessarily fully observable. In decentralized control problems we generally do not have full observability of the system. In the problems we consider, at best, the observations of all the agents combined determine with certainty the global state, and each such observation determines with certainty the partial view of each agent. The Meeting scenario is jointly fully observable and locally fully observable, but none of the agents know the complete state of the system.

Our next lemmas show some interesting relations between the classes identified so far. These lemmas will help us show in Section 3 that certain classes of Dec-POMDPs are easier to solve. The classes identified so far correspond to practical real-world scenarios such as multi-rover scenarios, multi-agent mapping and manufacturing where loosely-coupled robots act to achieve a global objective.

**Lemma 1** *If a Dec-MDP has independent observations and transitions, then the Dec-MDP is locally fully observable.*

**Proof.** We use properties of conditional probabilities established by Pearl (1988). The notation $I(X, Y|Z)$ means that the set of variables $X$ is independent of the set of variables $Y$ given the set of variables $Z$. The properties we use are:

- **Symmetry** — $I(X, Y|Z) \Leftrightarrow I(Y, X|Z)$.

- **Weak Union** — $I(X, Y \bigcup W|Z) \Rightarrow I(X, W|Z \bigcup Y) \ \wedge \ I(X, Y|Z \bigcup W)$.

- **Contraction** — $I(X, W|Z \bigcup Y) \ \wedge \ I(X, Y|Z) \Rightarrow I(X, Y \bigcup W|Z)$.

- **Decomposition** — $I(X, Y \bigcup W|Z) \Rightarrow I(X, Y|Z) \ \wedge \ I(X, W|Z)$.

Using this notation, the property of independent transitions implies $I(\{s_1'\}, \{s_2, a_2, s_2'\}|\{s_1, a_1\})$. The property of independent observations can be stated as $I(\{o_2\}, \{s_1, a_1, s_1', o_1\}|\{s_2, a_2, s_2'\})$.

From the independent observations and the weak union properties, we obtain

$$I(\{o_2\}, \{s_1'\}|\{s_2, a_2, s_2', s_1, a_1, o_1\}) \tag{1}$$

(where $X = \{o_2\}$, $W = \{s_1, a_1, o_1\}$, $Y = \{s_1'\}$ and $Z = \{s_2, a_2, s_2'\}$).

From the independent observations and decomposition property, we obtain:

$$I(\{o_1\}, \{s_2, a_2, s_2'\}|\{s_1, a_1, s_1'\}) \tag{2}$$

(where $X = \{o_1\}$, $Y = \{s_2, a_2, s_2'\}$, $W = \{o_2\}$ and $Z = \{s_1, a_1, s_1'\}$).

From the independent transitions, Equation 2 and the contraction properties, we obtain:

$$I(\{s_2, a_2, s_2'\}, \{s_1', o_1\}|\{s_1, a_1\}) \tag{3}$$

(where $X = \{s_2, a_2, s_2'\}$, $W = \{o_1\}$, $Z = \{s_1, a_1\}$ and $Y = \{s_1'\}$).

From Equation 3 and weak union we obtain:

$$I(\{s_2, a_2, s_2'\}, \{s_1'\}|\{s_1, a_1, o_1\}) \tag{4}$$

(where $X = \{s_2, a_2, s_2'\}$, $W = \{s_1'\}$, $Y = \{o_1\}$, and $Z = \{s_1, a_1\}$).

Applying the symmetry property to Equation 4, we obtain

$$I(\{s_1'\}, \{s_2, a_2, s_2'\} | \{s_1, a_1, o_1\}). \tag{5}$$

Applying the symmetry property to Equation 1 and contracting with Equation 5, we obtain

$$I(\{s_1'\}, \{s_2, a_2, s_2', o_2\} | \{s_1, a_1, o_1\}). \tag{6}$$

If we apply all of the above equations to the other agent (i.e., replace the index 2 with the index 1 and vice versa), then the following equation holds:

$$I(\{s_2'\}, \{s_1, a_1, s_1', o_1\} | \{s_2, a_2, o_2\}). \tag{7}$$

Applying the decomposition property to Equation 7, where $X = \{s_2'\}$, $W = \{s_1'\}$, $Y = \{s_1, a_1, o_1\}$ and $Z = \{s_2, a_2, o_2\}$ we obtain:

$$I(\{s_2'\}, \{s_1, a_1, o_1\} | \{s_2, a_2, o_2\}). \tag{8}$$

The lemma assumes a Dec-MDP, that is: $Pr(s'|o_1, o_2) = 1$. Since this probability is one, it is also true that $Pr(s'|s, a_1, a_2, o_1, o_2) = 1$. The lemma assumes independent transitions and observations, therefore the set of states is factored. Following conditional probabilities rules, we obtain:

$$1 = Pr(s_1', s_2' | s_1, a_1, o_1, s_2, a_2, o_2) = Pr(s_1' | s_1, a_1, o_1, s_2, a_2, s_2', o_2) Pr(s_2' | s_1, a_1, o_1, s_2, a_2, o_2).$$

Equation 6 means that $Pr(s_1' | s_1, a_1, o_1, s_2, a_2, s_2', o_2) = Pr(s_1' | s_1, a_1, o_1)$.
Equation 8 is equivalent to $Pr(s_2' | s_1, a_1, o_1, s_2, a_2, o_2) = Pr(s_2' | s_2, a_2, o_2)$.
Therefore,

$$1 = Pr(s_1', s_2' | s_1, a_1, o_1, s_2, a_2, o_2) = Pr(s_1' | s_1, a_1, o_1) Pr(s_2' | s_2, a_2, o_2).$$

So, each agent's partial view is determined with certainty by its observation and own transition, i.e., the Dec-MDP is locally fully observable. □

For the classes of problems studied in this lemma (Dec-MDPs with independent transitions and observations), a local policy for agent $i$ in a locally fully-observable Dec-MDP is a mapping from sequences of states in agent $i$'s partial view to actions. This differs from the general Dec-MDP case, where local policies are mappings from sequences of *observations* to actions. Formally, $\delta_i : S_i^* \rightarrow A_i$ where $S_i$ corresponds to the decomposition of global states assumed in Definitions 1 and 2 for Dec-MDPs with independent transitions and observations.

Moreover, we can show that an agent does not need to map a *sequence* of partial views to actions, but rather that it is sufficient to remember only the current partial view. This is shown in the next lemma.

**Lemma 2** *The current partial view of a state $s$ observed by agent $i$ ($s_i$) is a sufficient statistic for the past history of observations ($\overline{o}_i$) of a locally fully-observable Dec-MDP with independent transitions and observations.*

**Proof.** Without loss of generality we do all the computations for agent 1. We define $I_t^1$ as all the information about the Dec-MDP process available to agent 1 at the end of the control interval $t$. This is done similarly to Smallwood and Sondik's original proof for classical POMDPs (Smallwood & Sondik, 1973). $I_t^1$ is given by the action $a_{1_t}$ that agent 1 chose to perform at time $t$, the current resulting state $s_{1_t}$, which is fully observable by agent 1 ($s_{1_t} = i_1$), and the previous information $I_{t-1}^1$. We assume a certain policy for agent 2, $\pi_2$ is known and fixed. $\pi_2(s_t)$ is the action taken by agent 2 at the end of control interval $t$.

We compute the belief-state of agent 1, which is the probability that the system is at global state $j$ assuming only the information available to agent 1 ($I_t^1$). This computation allows us to build a belief-state MDP for agent 1. Agent 1's optimal local policy is the solution that obtains the highest value over all the solutions resulting from solving all the belief-state MDPs built for each possible policy for agent 2.

We compute the probability that the system is in state $s_t = j = (j_1, j_2)$ at time $t$, given the information available to agent 1: $Pr(s_t = j | I_t^1) = Pr(s_t = (j_1, j_2)| < a_{1_t}, s_{1_t}, \pi_2(s_t), I_{t-1}^1 >)$. Applying Bayes rule leads to the following result:

$$Pr(s_t = (j_1, j_2)| < a_{1_t}, s_{1_t}, \pi_2(s_t), I_{t-1}^1 >) = \frac{Pr(s_t = (j_1, j_2), s_{1_t} = i_1 | a_{1_t}, \pi_2(s_t), I_{t-1}^1)}{Pr(s_{1_t} = i_1 | a_{1_t}, \pi_2(s_t), I_{t-1}^1)}.$$

Since the Dec-MDP is locally fully observable, the denominator is equal to one. We expand the numerator by summing over all the possible states that could have lead to the current state $j$.

$$Pr(s_t = (j_1, j_2), s_{1_t} = i_1 | a_{1_t}, \pi_2(s_t), I_{t-1}^1) =$$

$$\Sigma_{s_{t-1}} Pr(s_{t-1}| a_{1_t}, \pi_2(s_t), I_{t-1}^1) Pr(s_t = j | s_{t-1}, a_{1_t}, \pi_2(s_t), I_{t-1}^1) Pr(s_{1_t} = i_1 | s_t = j, s_{t-1}, a_{1_t}, \pi_2(s_t), I_{t-1}^1)$$

The actions taken by the agents at time $t$ do not affect the state of the system at time $t-1$, therefore the first probability term is not conditioned on the values of the actions. The second probability term is exactly the transition probability of the Dec-MDP. Since the Dec-MDP has independent transitions, we can decompose the system transition probability into two corresponding probabilities $P_1$ and $P_2$, following Definition 1. The last term is equal to one because the Dec-MDP is locally fully observable. Therefore, we obtain:

$$Pr(s_t = j | I_t^1) = \Sigma_{s_{t-1}} Pr(s_{t-1}| I_{t-1}^1) P(s_t = j | s_{t-1}, a_{1_t}, \pi_2(s_t)) =$$

$$\Sigma_{s_{t-1}} Pr(s_{t-1}| I_{t-1}^1) P_1(s_{1_t} = j_1 | s_{1_{t-1}} = k_1, a_{1_t}) P_2(s_{2_t} = j_2 | s_{2_{t-1}} = k_2, \pi_2(s_t)).$$

Since agent 1 fully observes $s_1 = i_1$ at time $t$, then the probability that the system is at state $j$ and its first component $j_1$ is not $i_1$ is zero.

$$Pr(s_t = (j_1 \neq i_1, j_2) | I_t^1) = 0.$$

$$Pr(s_t = (i_1, j_2) | I_t^1) =$$

$$\Sigma_{s_{t-1}} Pr(s_{t-1}| I_{t-1}^1) P_1(s_{1_t} = i_1 | s_{1_{t-1}} = k_1, a_{1_t}) P_2(s_{2_t} = j_2 | s_{2_{t-1}} = k_2, \pi_2(s_t)).$$

Agent 1 can compute the last term for the fixed policy for agent 2. We conclude that the probability of the system being at state $j$ at time $t$ depends on the belief-state at time $t-1$. $\square$

Following this lemma, $s_i$, the current partial view of agent $i$ is a sufficient statistic for the history of observations. Therefore, an agent does not need to remember sequences of observations in order to decide which actions to perform.

**Corollary 1** *Agent $i$'s optimal local policy in a Dec-MDP with independent transitions and observations can be expressed as a mapping from agent $i$'s current partial view and current time to actions. For the finite horizon case:*

$$\delta_i : S_i \times T \to A_i.$$

The Meeting under Uncertainty scenario as described corresponds to a Dec-MDP with independent transitions and observations, therefore it is locally-fully observable. In such a case, for every possible location, a robot's decision about its next optimal move is not affected by the previous locations where the robot moved through.

We continue our classification of decentralized problems considering two additional dimensions: one is whether agents can share information and the other is whether the agents' behavior is goal-oriented. These classes are further described in the next two sections.

## 2.2 Information Sharing

We distinguish among three possible ways in which agents can share information: indirect communication, direct communication, and common uncontrollable features.

1. **Indirect Communication** — In the most general view, an action ($a_i \in A_i$) performed by an agent can result in three different consequences, and thus it serves any of the following three purposes: information gathering, changing the environment and indirect communication. Agent $i$'s actions can affect the observations made by agent $j$; these observations can serve as messages transmitted by agent $i$. Let's assume, for example, that a robot determines its location relative to the other robot's location. Then, the agents may have agreed on a meeting place based on their respective locations: If robot 1 sees robot 2 in location A, then they will meet at meeting place MA otherwise they will meet at meeting place MB. Even though the agents do not communicate directly, the dependencies between the observations can carry information that is shared by these agents.

   It should be noted that when assuming dependencies between observations that result in information sharing through indirect communication, the general decentralized control problem already includes the problem of what to communicate and when. That is, indirect communication is established as a consequence of an action performed by an agent and the observations sensed by the other agents as a result. Independent of the policy, this type of communication is limited to transferring only information about the features of the state. In a more general context, the meaning of the communication can be embedded in the policy. That is, each time that it makes an observation, each agent can infer what was meant by the communication in the domain and in the policy. This type of communication is assuming that the observations of the agents are indeed dependent and this dependency is actually the means that enables each agent to transmit information.

2. **Direct Communication** — Information can be shared by the agents if they can send messages directly to each other (e.g., robot 1 sends a message to robot 2: "Bring tool T to location (x,y)"). In this case, the observations can be either dependent or independent. We analyze decentralized processes with direct communication further in Section 5.

3. **Common Uncontrollable Features** — This is knowledge about environmental features that can be acquired by both agents but are not affected by any of these agents' actions. This common knowledge exists when there are features in the system state that are affected by the environment independent of the agents' actions. An example of such feature is the weather (assuming that neither of the agents can have any effect on whether it rains or it shines). Then, information about the weather can be made available to both agents if they share the same feature. Agents can then act upon the conditions of the weather and thus coordinate their actions without exchanging messages directly. They may have already decided that when the sun shines they meet at location MA, and otherwise at location MB.

Given that the global set of states $S$ is factored, a *common feature* $S^k$ is a feature of the global state that is included in the partial views of both agents.

**Definition 6 (Common Uncontrollable Features)** *A common feature is uncontrollable if:*

$$\forall a, b \in A_1, a \neq b, Pr(S^k|a, S) = P(S^k|b, S) \ \wedge \ \forall c, d \in A_2, c \neq d, Pr(S^k|c, S) = P(S^k|d, S).$$

It is important to note that the classes of problems that allow for uncontrollable features present an open problem that may be different from the categories studied in this paper due to the kind of dependencies that this knowledge may cause.

In this paper, we focus on either indirect communication or direct communication when we allow information sharing. We exclude from the discussion uncontrollable state features because this knowledge could provide a form of dependency between the agents that is beyond the scope of this paper.

**Assumption 1** *We assume that every change in the system results necessarily from the agents' actions.*[5]

Finally, the next section presents our last classification of decentralized problems that have goal-oriented behavior. This classification is practical in many areas where the agents' actions may incur some cost while trying to achieve a goal and may attain a global reward only when the goal is reached. This is different from most of the studies done on single-agent MDPs where a reward is obtained for every action performed.

---

5. Deterministic features that never change their values, or change their values in a deterministic way (such as time that increases in each step) are allowed.

### 2.3 Goal-oriented Behavior

We characterize decentralized processes in which the agents aim to reach specific global goal states. The Meeting under Uncertainty problem satisfies this requirement since the agents' goal is to meet at some location. Other practical scenarios may include assembling a machine, transferring objects from one location to a final destination, and answering to a query.

**Definition 7 (Finite-horizon Goal-oriented Dec-MDPs (GO-Dec-MDP))** *A finite-horizon Dec-MDP is* goal-oriented *if the following conditions hold:*

1. *There exists a special subset $G$ of $S$ of global goal states. At least one of the global goal states $g \in G$ is reachable by some joint policy.*

2. *The process ends at time $T$ (the finite horizon of the problem).*

3. *All actions in $A$ incur a cost, $C(a_i) < 0$. For simplicity, we assume in this paper that the cost of an action depends only on the action. In general, this cost may also depend on the state.*

4. *The global reward is $R(s, a_1, a_2, s') = C(a_1) + C(a_2)$.*

5. *If at time $T$, the system is in a state $s \in G$ there is an additional reward $JR(s) \in \Re$ that is awarded to the system for reaching a global goal state.*

A goal-oriented Dec-MDP has *uniform cost* when the costs of all actions are the same. When a goal oriented Dec-MDP has independent transitions and observations, we assume that there is a distinct action NOP with cost zero that has no effect on the local state and can only be performed at a global goal state component $g_i$. That is, NOP $\in A$ such that $C(\text{NOP}) = 0$ and $P_1(s'_1 | s_1, \text{NOP})$ is one when $s'_1 = s_1 = g^1_1$ for some global goal $g^i$.[6] When the GO-Dec-MDP has uniform cost and the set of actions include the NOP action, the uniform cost refers to all the actions different from NOP, i.e., $\forall a_i, a_j \in A \setminus \text{NOP}, C(a_i) = C(a_j)$.

Solving a GO-Dec-MDP is the problem of finding a joint policy that maximizes the global value. The definition is concerned with global goal-oriented behavior; it does not necessarily imply that each agent separately must achieve certain goals. In the next section, we analyze the complexity of various classes of Dec-POMDPs based on the above characterization.

## 3. A Taxonomy of Decentralized POMDPs: Complexity Results

We have distinguished between Dec-POMDPs and Dec-MDPs (where joint full observability is assumed). In neither case do the agents have full observability of the global state (at most they have *joint* full observability and *local* full observability, which are different from full observability). Therefore, each one of the agents has a belief about the global state of the system, represented as a probability distribution over global states. Table 1 presents the information that each agent needs in order to update its belief about the global state of the system. Since each agent can solve its belief-state MDP assuming a fixed and known policy for the other agent, the information required by an agent to update each belief-state reflects the complexity of solving each class of corresponding decentralized control problems. All the complexity results presented in this section apply to decentralized processes controlled by $n$ agents.

---

6. These NOP actions are necessary for agents that reach a component of a global goal state to "wait" for the other agent to complete its task.

| Process Class | Observations Needed by Agent $i$ | Reference |
|---|---|---|
| Dec-POMDP | The *local sequence* of observations: $\overline{o}_i$ | (Bernstein et al., 2002) |
| IT, IO Dec-POMDP | The *local sequence* of observations: $\overline{o}_i$ | Conjecture 1 |
| IT Dec-MDP (no IO) | The *local sequence* of observations: $\overline{o}_i$ | Conjecture 2 |
| IT, IO Dec-MDP | The *last local* observation: $o_i = s_i$ | Lemma 2 |

Table 1: A summary of the information upon which an optimal local policy is conditioned. IT stands for independent transitions and IO for independent observations.

**Conjecture 1** *An optimal policy of a Dec-POMDP with independent transitions and observations depends on the entire sequence of observations.*

**Conjecture 2** *An optimal policy of a Dec-MDP with independent transitions (but no independent observations) depends on the entire sequence of observations.*

It is an open question whether any belief-update scheme must memorize all the observations, for example, when there is partial observability because the process is not jointly fully observable or because the observations are dependent (see the first three cases in Table 1). It should be noted that only for the last case in the table have we shown that $s_i$ is a sufficient statistic (see Lemma 2).

This section studies to what extent the classes characterized in the previous section differ in complexity. In all the results shown below we refer to the complexity of finding an optimal joint policy for the decentralized control problems handled in the lemmas. The lemmas are stated for the corresponding decision problems (i.e., given the decentralized process assumed in each one of the lemmas, the decision problem is to decide whether there is a joint policy whose value is equal or larger than a given constant K). Nevertheless, finding a solution cannot be easier than deciding the same problem.

All the results in this section correspond to problems given with a finite horizon $T$. It is already known that deciding a finite-horizon decentralized MDP is NEXP-complete (Bernstein et al., 2002). In Section 2.2, we described indirect communication, i.e., situations in which information can be shared when the observations are dependent. Therefore, the same complexity result applies to the decentralized control problem with indirect-communication, as stated in the next corollary.

**Corollary 2** *Deciding a Dec-MDP or a Dec-POMDP with indirect communication (i.e., allowing the agents to communicate by acting and observing, when the observations are dependent) is NEXP-complete.*

We show in the next lemma that adding only goal-oriented behavior to a general decentralized process does not change the complexity of the problem. In other words, a goal-oriented decentralized problem is not easier than the general problem.

**Lemma 3** *Deciding a goal-oriented Dec-MDP is NEXP-complete.*

**Proof.** This case can be proved through the same reduction applied by Bernstein et al. (2002). We can reduce the general goal-oriented Dec-MDP problem to the tiling problem by adding a goal state to the last state of the Dec-MDP defined in the reduction. The agents reach this new goal state and receive a reward of zero if the tiling is consistent. Otherwise the agents obtain a reward of -1 and do not reach the goal state (but they do reach a terminal state and the process ends).

The main reason for this complexity result relies on the fact that each agent needs to remember a *sequence* of observations that it has sensed (see Table 1). Adding only goal states to the decentralized process (without assuming any further assumptions) does not make the control problem any easier. □

Since a Dec-POMDP is more general than a Dec-MDP, the same lower bound for the Dec-MDP is valid.

**Corollary 3** *Deciding a goal-oriented Dec-POMDP is NEXP-complete.*

The next lemma shows that by assuming that the transitions and observations are independent and that the agents have joint full observability, the problem of solving a decentralized cooperative system becomes easier (the lemma does not assume goal-oriented behavior).

**Lemma 4** *Deciding a Dec-MDP with independent transitions and observations is NP-complete.*

**Proof.** The Dec-MDP is locally fully observable because it has independent transitions and observations (see Lemma 1). We have shown in Lemma 2 that for such Dec-MDPs, the current partial view of an agent is a sufficient statistic. Therefore, a local policy of an agent $i$ is of size polynomial in $|S_i|T$.[7] There are $|A_i|^{|S_i|T}$ policies (mappings from $S_i$ and time to $A_i$). Each agent $i$ needs to build a belief-state MDP with a number of states that is polynomial in $|S_i|$ (for a fixed and known policy for the other agent). Evaluating one such local policy can be done in polynomial time (by running dynamic programming on the belief-state MDP), but there is an exponential number of such policies for which this should be done. Therefore, the upper bound for the decision problem stated in this lemma is NP.

Figure 2 shows schematically the differences in the policies, which lead to the difference in the complexity classes of the control problems when independent transitions and observations are assumed. When no assumptions are made (as in the leftmost figure), a local policy is represented by a tree, where each node corresponds to a possible action and each edge corresponds to a possible observation (i.e., a possible transition). In each local policy, the agent needs to remember a sequence of observations as opposed to just the last observation as in the rightmost figure. In the belief-state MDP that each agent builds, there is an exponential number of states that correspond to all the possible sequences of observations (this number is $|\Omega_i|^T$, $T$ is the finite horizon). Each such policy (of exponential size) can be evaluated with dynamic programming. There are a total of $|A_i|^{|\Omega_i|^T}$ local policies for which such a belief-state MDP need to be built by a brute-force search algorithm.[8]

---

7. We proved that $s_i$ is a sufficient statistic, i.e., the current local state summarizes all the observations seen so far. However, since we are studying finite-horizon problems, a local policy of action is not stationary. Therefore, the time is indeed needed to decide upon an action.

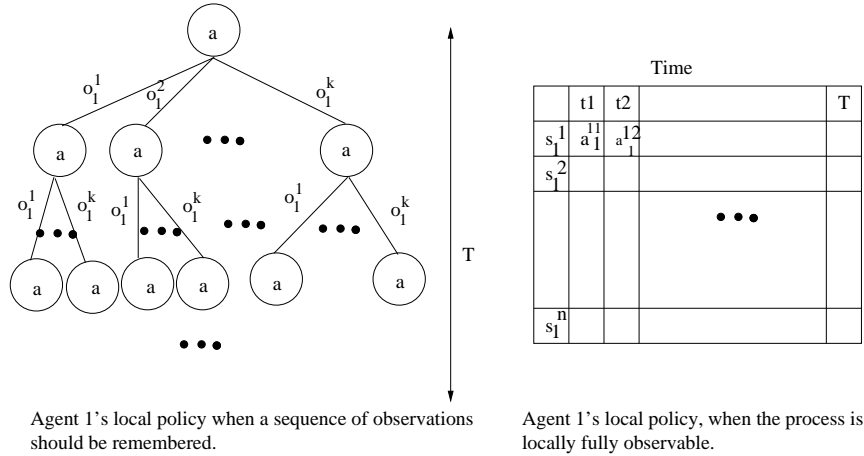

Agent 1's local policy when a sequence of observations should be remembered.

Agent 1's local policy, when the process is locally fully observable.

Figure 2: Exponential vs. Polynomial Sized Policies.

It is already known that a simple decentralized decision-making problem for two agents is NP-hard (where $|A_i| \geq 2$ and $|A_j| \geq 3$) (Papadimitriou & Tsitsiklis, 1982, 1986). Therefore, the lower bound for the problem class stated in the lemma is also NP. □

It is an open question whether a Dec-POMDP with independent transitions and observations (without joint full observability) results in a complexity class lower than NEXP.

Easier problems are obtained when the class of decentralized problems is restricted to goal-oriented behavior and does not include any type of information sharing, while maintaining the independence assumptions.

**Lemma 5** *Deciding a goal-oriented Dec-MDP with independent transitions and observations, with a single global goal state and with uniform cost is P-complete.*

**Proof.** We argue that each agent should follow locally the policy that minimizes the cost to $g_i$ by solving a single-agent MDP. Under the uniform cost assumption, this is equivalent to minimizing the number of steps. Because computing these policies can be done with dynamic programming, the problem is P-complete.

In a given local state, in general, agents could follow the shortest path to $g_i$ or abandon it and do something more beneficial. Abandoning the goal is never beneficial because every course of action is equally valuable (there are no intermediate rewards and all actions' costs are the same). □

---

8. Assuming that $T$ is similar in size to $|S|$, we obtain that the complexity of the brute-force search algorithm is double exponential in $|S|$. If $T << |S|$ the complexity can be NP.

We extend this result to the case involving multiple global goal states. Unfortunately, it is not possible to avoid the need to change local goal states even when no information exchange is possible in GO-Dec-MDPs with independent transitions and observations. However, we will provide a necessary and sufficient condition for agents so that the need to change a goal never arises. Under this condition, the best strategy is to minimize the cost of reaching the local components of the single most beneficial goal. The analysis that we perform is only for the case of uniform cost of actions (i.e., all actions different from NOP incur the same cost). Without this assumption, it is sometimes beneficial to "waste time" and move away from goals because of the high cost/low likelihood to reach a goal.

Let $\delta_1^i$ be the best local policy for agent 1 to reach $g_1^i$ (computed as if $g^i$ is the only goal). Similarly, we define $\delta_2^i$ for agent 2 to reach its corresponding component of global goal $g^i$.

Let $\alpha_{i/j}^{s_1}$ be the probability of reaching $g_1^j$ from some state $s_1$, when executing the policy $\delta_1^i$. $s_1$ is some state reachable by policy $\delta_1^i$. Similarly, let $\beta_{i/j}^{s_2}$ be the corresponding probability for agent 2.

Let $\overline{C}(s_1, \delta_1^i)$ be the expected cost incurred by agent 1 when it executes policy $\delta_1^i$ starting from state $s_1$. Similarly, $\overline{C}(s_2, \delta_2^i)$ is defined for agent 2.

**Property 1 (No Benefit to Change Local Goals)** *Let $g^i$ be a global goal state such that the joint policy $(\delta_1^i, \delta_2^i)$ has the highest value over all global goal states.*

*A GO-Dec-MDP satisfies the NBCLG property if and only if the following two conditions hold:*

1. *$\forall j \neq i, s_1$ reachable from $s_1^0$ by $\delta_1^i$,*

   $$\Sigma_k \alpha_{i/k}^{s_1} \beta_{i/k}^{s_1^0} JR(g^k) + \overline{C}(s_1, \delta_1^i) + \overline{C}(s_2^0, \delta_2^i) \geq \Sigma_k \alpha_{j/k}^{s_1} \beta_{i/k}^{s_1^0} JR(g^k) + \overline{C}(s_1, \delta_1^j) + \overline{C}(s_2^0, \delta_2^i).$$

2. *$\forall j \neq i, s_2$ reachable from $s_2^0$ by $\delta_2^i$,*

   $$\Sigma_k \alpha_{i/k}^{s_2^0} \beta_{i/k}^{s_2} JR(g^k) + \overline{C}(s_1^0, \delta_1^i) + \overline{C}(s_2, \delta_2^i) \geq \Sigma_k \alpha_{i/k}^{s_2^0} \beta_{j/k}^{s_2} JR(g^k) + \overline{C}(s_1^0, \delta_1^i) + \overline{C}(s_2, \delta_2^j).$$

Note that the property guarantees that whenever an agent is at an intermediate local state, if the agent reevaluates the value of continuing to optimize its path to its goal $i$ versus switching to another goal $j$, the value of continuing the policy to the original goal always remains the highest. Hence, it is never beneficial to change local goals. When the condition is not met, however, it is beneficial to change a local goal despite the fact that there is no information exchange between the agents.

**Lemma 6** *Deciding a goal-oriented Dec-MDP with independent transitions and observations with at least one global goal state and with uniform cost is P-complete when the NBCLG property is satisfied.*

**Sketch of Proof.** For each global goal state $g^k \in G$, agent 1 can compute its optimal local policy $\delta_1^k$ to reach the local component $g_1^k$. Similarly, agent 2 can compute its optimal local policy $\delta_2^k$ for every goal $g^k$. This is achieved by solving single agent MDPs, each aimed at reaching a local component of a global goal state: $MDP_1^k = <S_1, A_1, P_1, R_1^k>$ and $MDP_2^k = <S_2, A_2, P_2, R_2^k>$. $S_1$ and $S_2$ are given from the factored representation of the global states of the system. $P_1$ and $P_2$ result from the independent transitions assumption. The local reward $R_1^k(s_1, a_1, s_1')$ is the sum of the cost of an action, $C(a_1)$ and an additional arbitrary reward attained at time $T$ if the state $s_1'$ reached at $T$ is $g_1^k$. $R_2$ is defined similarly for agent 2. We denote by $g^i$ the global goal state that attains the highest value when the agents follow the joint policy $(\delta_1^i, \delta_2^i)$.

The NBCLG property implies that when agent 1 executes its optimal local policy $\delta_1^i$, it cannot reach a state in which it is beneficial to abandon $g_1^i$ and pursue a different local component $g_1^j$. (Without the property, this situation may arise despite the fact that agent 2 may continue to pursue $g_2^i$.) Note that this remains true even though the local policy $\delta_1^i$ may bring agent 1 eventually to a component of a goal that is different from the goal $i$. Since this property includes the same condition for both agents, it is also true that agent 2 will have no incentive to switch from an optimal local policy to goal $g^i$, $\delta_2^i$, once it starts executing it at the initial state. Therefore, the joint policy $(\delta_1^i, \delta_2^i)$ is the optimal solution to the given GO-Dec-MDP. Since computing $\delta_1^i$ and $\delta_2^i$ respectively can be done with dynamic programming, a GO-Dec-MDP with independent transitions and observations and with uniform cost that satisfies the NBCLG is P-complete. $\square$

Assume that the Meeting scenario is given with a single meeting location (e.g., the lander in the rovers' case). Then, given the other independence assumptions, we can solve this problem optimally by building single-agent MDPs, each designed to achieve its component of the goal state (in the example, each robot needs to reach the lander). If there is a larger set of global goal states (e.g., when there is a finite number of possible meeting sites such as the lander, the space station on the planet and some other particular site) then following the lemma, we can let each agent find its local optimal policy to its corresponding component in these goal states. The optimal joint policy is the pair of local policies with the highest value.

A summary of the complexity results presented in this section appears in Figure 3.

## 4. Algorithms for Decentralized Control with No Information Sharing

So far, the only known algorithms for solving optimally decentralized control problems are the generalized version of dynamic programming for Dec-POMDPs (Hansen et al., 2004) and the Coverage-set algorithm (Becker et al., 2003) for Dec-MDPs with independent transitions and observations. The first algorithm solves optimally a general Dec-POMDP. Its practicality is restricted by the complexity of these problems (NEXP-complete). The Coverage-set algorithm assumes that the agents' actions could result in super-additive or sub-additive joint rewards as follows. In the first case, the reward obtained by the system from agents doing certain actions is larger than the sum of each agent's local reward for those actions. In the second case, sub-additive joint rewards will be attained when the agents are penalized for doing redundant actions. As an example, we can look at a modified version of the Meeting scenario, where robots can move and also run experiments at different sites. Then, a process may lead to sub-additive rewards if both agents run the same experiment, wasting their resources instead of doing non-overlapping tasks. In other cases, the system may benefit when both robots perform the same tasks. For example, both agents run the same experiment at different times in the day, collecting eventually results with better quality. The class of problems handled by the Coverage-set algorithm does not include necessarily goal-oriented decentralized processes. In this section, we present two tractable algorithms for controlling optimally Dec-MDPs with independent transitions and observations, which are also goal-oriented. A summary of the algorithms that are known to solve decentralized control problems optimally is presented in Table 2.

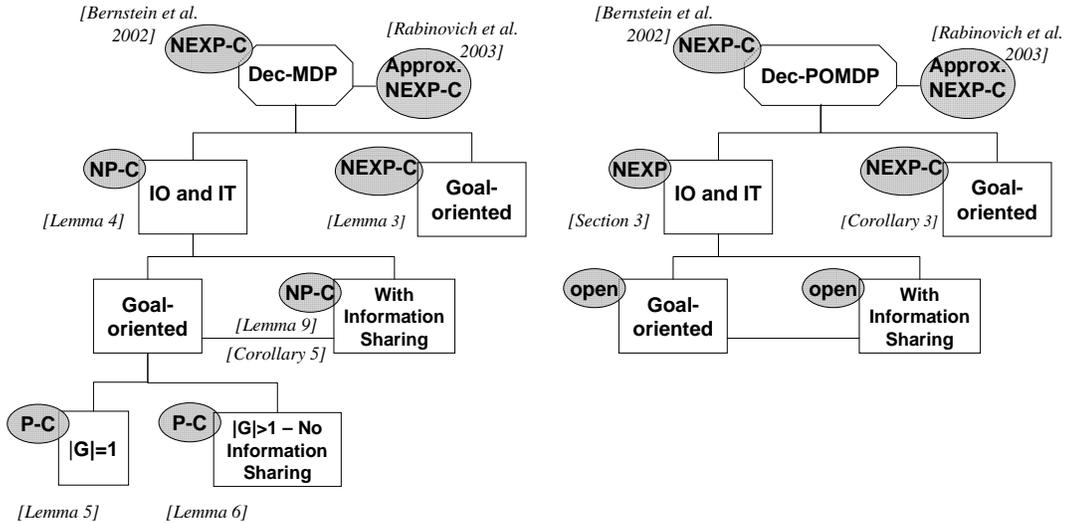

Figure 3: A summary of the complexity analysis for classes of decentralized control processes. We use the notation IT and IO for independent transitions and observations respectively.

## 4.1 Single-goal, Goal-oriented Dec-MDPs

Following Lemma 5, the optimal solution for a GO-Dec-MDP with a single global goal state is computed by solving single agent MDPs aimed at the corresponding components of the given global goal state. Algorithm $Opt1Goal$ is shown in Figure 5. Each agent $i$'s MDP finds the least cost path to $g_i$. Because the cost of actions (different from NOP) is uniform any value set to $GR(g_i)$ is suitable to solve the problems.

## 4.2 Many-goals, Goal-oriented Dec-MDPs

Due to the uncertainty about the outcomes of actions, an agent may decide to change its intention with respect to the global goal state it is planning to reach. The algorithm that optimally and decentrally solves a goal-oriented Dec-MDP problem with many global goal states is $OptNGoals$, presented in Figure 5. Assuming the conditions of Lemma 6, this algorithm is optimal because the lemma guarantees that changing local goals is not beneficial.

---

9. No algorithm was proposed short of full search with complexity NP as shown in Lemma 9.

| Process Class | Optimal Algorithm | Reference |
|---|---|---|
| Dec-POMDP | Gen. dynamic programming | (Hansen et al., 2004) |
| IT, IO Dec-MDP, no information sharing | Coverage-set | (Becker et al., 2003) |
| IT, IO Dec-MDP, with direct communication | Not Known Yet[9] | Section 5 |
| IT, IO, GO-Dec-MDP ($|G|=1$) no information sharing, with uniform cost | *Opt1Goal* | Section 4.1 |
| IT, IO GO-Dec-MDP ($|G| \geq 1$), no information sharing with uniform cost, NBCLG property | *OptNGoals* | Section 4.2 |
| IT, IO GO-Dec-MDP ($|G| \geq 1$), with direct communication | Not Known Yet[9] Treated in (Goldman & Zilberstein, 2004). | Section 5 |

Table 2: A summary of the known algorithms for controlling decentralized MDPs optimally.

---

**function** Opt1Goal(Dec-MDP)
    **returns** the optimal joint policy $\delta^*$,
    **inputs:**   Dec-MDP=$<S, A_1, A_2, P, R>$
           $G$   /* the set of global goal states, $|G|=1$, $g=(g_1, g_2) \in G \subseteq S^*$/
           $/*$ *Transition independence* $\Rightarrow S=S_1 \times S_2, P=P_1 \times P_2$ */
           $/* R(s, a_1, a_2, s') = Cost(a_1) + Cost(a_2) + \text{JR}(s') * /$
           /*ComputeLocalR computes the local rewards as follows:*/
           $/* R_i(s_i, a_i, s_i') = Cost(a_i) + \text{GR}(s_i') * /$
           $/* \text{GR}(s_i') \in \Re$ *if* $s_i'=g_i$, *else* $0 * /$

    $R_1 \leftarrow ComputeLocalR(S_1, A_1, P_1, g_1)$
    $MDP_1 = <S_1, A_1, P_1, R_1>$
    $R_2 \leftarrow ComputeLocalR(S_2, A_2, P_2, g_2)$
    $MDP_2 = <S_2, A_2, P_2, R_2>$
    $\delta_1^* \leftarrow SOLVE(MDP_1)$
    $\delta_2^* \leftarrow SOLVE(MDP_2)$
    $\delta^* \leftarrow (\delta_1^*, \delta_2^*)$
    **return** $\delta^*$

Figure 4: The Opt1Goal Algorithm.

Following the *OptNGoals* algorithm, each agent solves iteratively its induced MDP towards each one of the possible components of each one of the global goal states. Finally, the optimal joint policy is the one with the highest value. This algorithm is described for a situation where the global goal states are distinctive, that is

$$\forall s_1^1, s_1^2 \in S_1, s_2^1, s_2^2 \in S_2 : (s_1^1, s_2^1), (s_1^2, s_2^2) \in G \Rightarrow (s_1^1 = s_1^2 \Leftrightarrow s_2^1 = s_2^2).$$

It is possible to extend the algorithm to non-distinctive global goal states, by running the algorithm over all possible subsets of goals $G_1, G_2$ such that $\forall s_1 \in G_1, s_2 \in G_2 : (s_1, s_2) \in G$.

---

**function** OptNGoals(Dec-MDP)
    **returns** the optimal joint policy $\delta^*$,
    **inputs:**    Dec-MDP=$<S, A_1, A_2, P, R>$
              $G$  /* the set of global goal states, $|G| = N, g_i = (g_1^i, g_2^i) \in G, 1 \le i \le N */$
             /* *Transition independence* $\Rightarrow S = S_1 \times S_2, P = P_1 \times P_2 */$
             /* $R(s, a_1, a_2, s') = Cost(a_1) + Cost(a_2) + \text{JR}(s') */$
             /* $R_i(s_i, a_i, s_i') = Cost(a_i) + \text{GR}(s_i') */$
             /* $\text{GR}(s_i') \in \Re \ if \ s_i' = g_i, \ else \ 0 */$

    $CurrOptJoint\delta \leftarrow Opt1Goal(\text{Dec-MDP}, (g_1^1, g_2^1))$
    $CurrMaxVal \leftarrow ComputeV(\text{Dec-MDP}, CurrOptJoint\delta, s^0)$
    **for** $i \leftarrow 2$ **to** $N$
        $\delta^{*i} \leftarrow Opt1Goal(\text{Dec-MDP}, (g_1^i, g_2^i))$
        $CurrVal \leftarrow ComputeV(\text{Dec-MDP}, \delta^{*i}, s^0)$
        **if** $(CurrVal > CurrMaxVal)$ **then**
            $CurrOptJoint\delta \leftarrow \delta^{*i}$
            $CurrMaxVal \leftarrow CurrVal$
    **return** $CurrOptJoint\delta$

**function** ComputeV (Dec-MDP,$\delta, s^0$)
    **returns** the value of state $s^0$ following joint policy $\delta$, $V_\delta(s^0)$
    **inputs:**    Dec-MDP,
              $\delta = (\delta_1, \delta_2)$, the joint policy found so far .
              $s^0$, the initial state of the Dec-MDP.

    Run Value Iteration to compute $V^\delta(s)$ for all $s \in S$:
    $V^\delta(s) = \Sigma_{s'=(s_1', s_2')} P_1(s_1'|\delta_1(s_1), s_1) P_2(s_2'|\delta_2(s_2), s_2)(R(s, \delta_1(s_1), \delta_2(s_2), s') + V^\delta(s'))$
    **return** $V^\delta(s^0)$

---

Figure 5: The OptNGoals Algorithm.

## 5. Decentralized Control with Communication

Direct communication can be beneficial in decentralized control because the agents lack full observability of the global state. That is, the value of the optimal joint policy that allows communication may be larger than the value of the optimal joint policy without communication. We are interested in solving a decentralized control problem off-line taking into account direct communication. Agents will consider this expected information while computing their optimal joint policy, thus deriving a policy for when and what to communicate.

If we assume that direct communication leads to full observability of the system state that direct communication is free and that the observations are independent then obviously the agents will benefit most by communicating constantly. This results in a fully observable decentralized process, which is equivalent to an MMDP (Boutilier, 1999). This problem is known to be P-complete (Papadimitriou & Tsitsiklis, 1987).

In real-world scenarios, it is reasonable to assume that direct communication has indeed an additional cost associated with it; the cost may reflect the risk of revealing information to competitive agents, the bandwidth necessary for the transmission or even the complexity of computing the information to be transferred. Therefore, communication may not be possible or even desirable at all times.

We extend the model of decentralized partially-observable Markov decision process to include an explicit language of communication with an associated cost.[10] We call this model Dec-POMDP-Com. It is given by the following tuple: $< S, A_1, A_2, \Sigma, C_\Sigma, P, R, \Omega_1, \Omega_2, O, T >$.

$\Sigma$ denotes the alphabet of messages and $\sigma_i \in \Sigma$ represents an atomic message sent by agent $i$ (i.e., $\sigma_i$ is a letter in the language). $\overline{\sigma_i}$ denotes a sequence of atomic messages. A special message that belongs to $\Sigma$ is the null message, which is denoted by $\epsilon_\sigma$. This message is sent by an agent that does not want to transmit anything to the other agents.[11] $C_\Sigma$ is the cost of transmitting an atomic message: $C_\Sigma : \Sigma \rightarrow \Re$. The cost of transmitting a null message is zero. Communication cost models determine the flow of the information exchange and the cost of this communication. These models may include, for example, one-way communication models, in which the cost $C_\Sigma$ is incurred by each agent that sends information to another agent and two-way communication models, where agents exchange messages when at least one of them initiates communication, and the cost is incurred only once each time (we refer to these agents as jointly exchanging messages). Other models may require additional messages like acknowledgments that may incur additional costs.

We restrict ourselves in this paper to communication cost models based on joint exchange of messages and to communication that leads to full observability of the global state. The agents send messages by broadcasting, and only one message is sent at each time. The agents in the system share the same language of communication. In a separate line of research, we are addressing the question of agents controlling a decentralized process where the agents develop a mutual understanding of the messages exchanged along the process (Goldman, Allen, & Zilberstein, 2004). Direct communication is the *only* means of achieving full observability when the observations are independent and when there is no common uncontrollable features (Assumption 1).

We define a Dec-MDP-Com as a Dec-POMDP-Com with *joint* full observability, as we did with Dec-POMDPs and Dec-MDPs in Section 2.1. The Dec-POMDP-Com model can have independent transitions, independent observations, be locally fully observable, and goal-oriented as the basic model presented in Section 2.

---

10. The model was presented by Goldman and Zilberstein (2003).
11. We omit in this paper the details of the communication network that may be necessary to implement the transmission of the messages.

We describe the interaction among the agents as a process in which agents perform an action, then observe their environment, and then send a message that is instantaneously received by the other agent.[12] Then, we can define the local policies of the controlling agents as well as the resulting joint policy whose value we are interested in optimizing. A *local policy* $\delta_i$ is composed of two policies: $\delta_i^A$ that determines the actions of the agents and $\delta_i^\Sigma$ that states the communication policy. Notice that $\delta_i^A$ allows indirect communication if the observations of the agents are dependent and that $\delta_i^\Sigma$ allows direct communication even when the observations are dependent.

**Definition 8 (Local Policy for Action)** *A local policy for action for agent $i$, $\delta_i^A$, is a mapping from local histories of observations $\overline{o_i} = o_{i_1}, \ldots, o_{i_t}$ over $\Omega_i$ and histories of messages $\overline{\sigma_j} = \sigma_{j_1}, \ldots, \sigma_{j_t}$,[13] received $(j \neq i)$ since the last time the agents exchanged information to actions in $A_i$.*

$$\delta_i^A : S \times \Omega^* \times \Sigma^* \to A_i.$$

**Definition 9 (Local Policy for Communication)** *A local policy for communication for agent $i$, $\delta_i^\Sigma$, is a mapping from local histories of observations $\overline{o_i} = o_{i_1}, \ldots, o_{i_t}$ and $o$, the last observation perceived after performing the last local action, over $\Omega_i$ and histories of messages $\overline{\sigma_j} = \sigma_{j_1}, \ldots, \sigma_{j_t}$, received $(j \neq i)$ since the last time the agents exchanged information to messages in $\Sigma$.*

$$\delta_i^\Sigma : S \times \Omega^* o \times \Sigma^* \to \Sigma.$$

More complex cases result if the agents could communicate partial information about their partial views. This is left for future work.

**Definition 10 (Joint Policy)** *A joint policy $\delta = (\delta_1, \delta_2)$ is defined as a pair of local policies, one for each agent, where each $\delta_i$ is composed of the communication and the action policies for agent $i$.*

The optimal joint policy that stipulates for each decision-maker how it should behave and when it should communicate with other agents is the policy that maximizes the value of the initial state of the Dec-POMDP-Com. We first study the general problem, when no particular assumptions are made on the class of the Dec-POMDP-Com. We will then study certain classes of this problem as we did with the case without communication.

**Definition 11 (Transition Probability Over a Sequence of States)** *The probability of transitioning from a state $s$ to a state $s'$ following the joint policy $\delta = (\delta_1, \delta_2)$ while agent 1 sees observation sequence $\overline{o_1} o_1$ and receives sequences of messages $\overline{\sigma_2}$, and agent 2 sees $\overline{o_2} o_2$ and receives $\overline{\sigma_1}$ of the same length, written $\overline{P_\delta}(s'|s, \overline{o_1} o_1, \overline{\sigma_2}, \overline{o_2} o_2, \overline{\sigma_1})$ can be defined recursively:*

*1. $\overline{P_\delta}(s|s, \epsilon, \epsilon, \epsilon, \epsilon) = 1$.*

---

12. When agents exchange information there is a question whether information is obtained instantaneously or there are delays. For simplicity of exposition we assume no delays in the system.

13. The notation $\overline{o} = o_1, \ldots, o_t$ and $\overline{o}o$ represents the sequence $o_1, \ldots, o_t o$. Similarly, the notation for sequences of messages: $\overline{\sigma_i}\sigma$ represents the sequence $\sigma_{i_1}, \ldots, \sigma_{i_t}\sigma$.

2. $\overline{P_\delta}(s'|s, \overline{\sigma_1}o_1, \overline{\sigma_2}\sigma_2, \overline{\sigma_2}o_2, \overline{\sigma_1}\sigma_1) =$

$$\sum_{q \in S} \overline{P_\delta}(q|s, \overline{\sigma_1}, \overline{\sigma_2}, \overline{\sigma_2}, \overline{\sigma_1}) * P(s'|q, \delta_1^A(s, \overline{\sigma_1}, \overline{\sigma_2}), \delta_2^A(s, \overline{\sigma_2}, \overline{\sigma_1})) *$$

$$O(o_1, o_2|q, \delta_1^A(s, \overline{\sigma_1}, \overline{\sigma_2}), \delta_2^A(s, \overline{\sigma_2}, \overline{\sigma_1}), s')$$

such that $\delta_1^\Sigma(s, \overline{\sigma_1}o_1, \overline{\sigma_2}) = \sigma_1$ and $\delta_2^\Sigma(s, \overline{\sigma_2}o_2, \overline{\sigma_1}) = \sigma_2$.

Then, the value of the initial state $s^0$ of the Dec-POMDP-Com after following a joint policy $\delta$ for $T$ steps can be defined as follows:

**Definition 12 (Value of an Initial State Given a Policy)**  *The value $V_\delta^T(s^0)$ after following policy $\delta = (\delta_1, \delta_2)$ from state $s^0$ for $T$ steps is given by:*

$$V_\delta^T(s^0) = \sum_{(\overline{\sigma_1}o_1, \overline{\sigma_2}o_2)} \sum_{q \in S} \sum_{s' \in S} \overline{P_\delta}(q|s^0, \overline{\sigma_1}, \overline{\sigma_2}, \overline{\sigma_2}, \overline{\sigma_1}) * P(s'|q, \delta_1^A(s^0, \overline{\sigma_1}, \overline{\sigma_2}), \delta_2^A(s^0, \overline{\sigma_2}, \overline{\sigma_1})) *$$

$$R(q, \delta_1^A(s^0, \overline{\sigma_1}, \overline{\sigma_2}), \delta_1^\Sigma(s^0, \overline{\sigma_1}o_1, \overline{\sigma_2}), \delta_2^A(s^0, \overline{\sigma_2}, \overline{\sigma_1}), \delta_2^\Sigma(s^0, \overline{\sigma_2}o_2, \overline{\sigma_1}), s')$$

*where the observation and the message sequences are of length at most $T - 1$, and both sequences of observations are of the same length $l$. The sequences of messages are of length $l + 1$ because they considered the last observation resulting from the control action prior to communicating.*

The problem of decentralized control with direct communication is to find an optimal joint policy $\delta^*$ for action and for communication such that $\delta^* = argmax_\delta V_\delta^T(s^0)$.

## 5.1 Languages of Communication

We start showing that under some circumstances the language of observations is as good as any other communication language. In the Meeting scenario, no matter what are the tasks assigned to the system, agents that exchange their current coordinates are guaranteed to find the optimal solution to the decentralized problem.

**Theorem 1**  *Given a Dec-MDP-Com with constant message cost, the value of the optimal joint policy $\delta^*$ with respect to any $\Sigma$, $V_{\delta^*,\Sigma}^T(s^0)$ is not greater than the value of the optimal joint policy with respect to the language of observations ($\Sigma = \Omega$). That is:*

$$\forall \Sigma \; V_{\delta^*,\Sigma}^T(s^0) \leq V_{\delta^*,\Sigma=\Omega}^T(s^0).$$

**Proof.**  The decentralized process is jointly fully observable. Therefore, it is not beneficial for the agents to send any information in addition to their observations (we assume joint exchange of messages). Thus, combining both agents' observations results in the complete global state. Moreover, the theorem assumes a constant cost for every message, i.e., all non-null messages incur the same cost: there are not any messages that are either more expensive or cheaper to transmit than others. Therefore, the agents cannot benefit from exchanging information that is a strict subset of their partial views because the cost of sending any message is equal. $\square$

We note that the theorem does not hold when different messages may incur different costs. In this case, sending less information might be cheaper, but equally valuable. For example, when agents observe their respective $x$ and $y$ coordinates, they may benefit from sending only one coordinate if it costs less than sending the complete location. Agents may also benefit from sending functions of their observations if this incurs a smaller cost. For example, agents may benefit from exchanging information about the Manhattan distance between their current location and some mutually-known location.

In general, it seems reasonable to introduce a language of communication to reduce complexity, but as the theorem shows, this cannot guarantee optimality when the language is comprised of messages different from the agents' observations. Examples of such messages include: 1) commitments, which are constraints on the future behavior of the message sender, 2) instructions, which are constraints on the future behavior of the message receiver, and 3) feedback that is an encouraging or punishing signal that is sent to another agent. The study of Dec-POMDP-Com problems with languages of communication different from observations is left for future work. Similarly, certain protocols of communication can restrict the optimal value of the policy of communication but may be easier to implement.

## 5.2 The Effect of Communicating on the Complexity Analysis

The complexity results we obtained in Section 3 apply also for the same classes of problems when direct communication is possible. Although agents achieve full observability each time they exchange information, the problem of finding the policy of communication off-line (when there is a cost associated with each communication act) remains as hard as the general problem with no communication. In the worst case, transmitting the messages can be prohibitively expensive. Therefore, adding direct communication does not simplify the problem. For all the cases shown to be in NEXP, adding direct communication cannot make them more difficult. The complexity of deciding a Dec-MDP when observations are independent and direct communication is allowed remains the same as when direct communication is not assumed, as shown in Lemma 9. The impact of direct communication on the classes of Dec-POMDPs with independent transitions and observations and with possible goal-oriented behavior remains an open question.

It is interesting to note that the decentralized control problem with direct communication can be reduced to the same problem with indirect communication when the observations are dependent. We assume that transmitting messages incur the same cost and that the language of messages is the language of observations. If the language of communication is different then the reduction does not apply.

**Lemma 7** *A Dec-MDP with direct communication is polynomially-reducible to a Dec-MDP with indirect communication.*

**Proof.** We denote the Dec-MDP with direct communication as Dec-D, and the Dec-MDP with indirect communication as Dec-I. The reduction from Dec-D to Dec-I requires the addition of a single bit $b$ to the global states of Dec-I. When $b$ takes the value 1, the agents are in the communication mode. When $b$ takes the value 0, the agents are performing control actions. A communication action $a^c$ performed by agent $i$ is agent $i$'s local observation $o_i$. The transition probability of Dec-I, $P_I$ is given as follows: $P_I([s, 1], o_1, o_2, [s, 0]) = 1$, no change is caused to the global state of the system besides flipping the value of $b$ back to 0 each time the agents exchange information. The probability of observing $o_1$ and $o_2$ (respectively by the two agents) after performing communication acts when $b$ equals 1 is one as long as $o_1$ is agent 2's last observation, and $o_2$ is agent 1's last observation. This probability is zero for any other action taken at $[s, 1]$. $O(o_2, o_1 | [s, 1], o_1, o_2, [s, 0]) = 1$. $\square$

Theorem 1 showed that exchanging observations is sufficient to guarantee that the joint policy computed will be optimal. The next lemma shows that the optimal policy of communication will instruct the agent to transmit only its current observation or the null message.

**Lemma 8** *Given a Dec-MDP-Com with constant message cost, there is an optimal policy of communication such that whenever a non-null message is sent, it must be the agent's last observation.*

**Proof.** This lemma results from Theorem 1. In a jointly-fully observable process, sending a non-null message that is an observation different from the last one cannot provide more information about the current state of the process than the last observation does. $\square$

Since the current global state becomes fully observable each time that the agents communicate, all the necessary information is stored in the synchronized state $s$; the agents do not need to remember all the messages received so far when the decentralized process is jointly fully observable. When the problem is framed as a Dec-MDP-Com with independent transitions and observations, the current local state is fully observable. Thus, the local policies of action and communication for such a Dec-MDP-Com can be formalized as follows:

**Corollary 4** *An optimal local policy of action for this problem, $\delta_i^A$, can be represented as a mapping from synchronized states, current partial views, and time to actions.*

$$\delta_i^A : S \times S_i \times T \to A_i.$$

*Similarly, an optimal local policy of communication $\delta_i^\Sigma$ can be represented as a mapping from synchronized states, current partial views, and time to two possible messages: either the current partial view or the null message.*

$$\delta_i^\Sigma : S \times S_i \times T \to S_i \cup \{\epsilon_\sigma\}$$

*such that if $\delta_i^\Sigma(s, s_i) \neq \epsilon_\sigma$ then $\delta_i^\Sigma(s, s_i) = s_i$.*

Agents need to remember only their current partial view and the last synchronized information to decide on their next action. This is a primary observation that affects the complexity of deciding Dec-MDP-Com with independent transitions and observations as shown in the next lemma.

**Lemma 9** *Deciding a Dec-MDP-Com with independent transitions and observations is NP-complete.*

**Proof.** Following Corollary 4, each agent's policy is of size polynomial in $|S|$, and the number of possible policies is $2^{|S|^2 T} \times |A|^{|S|^2 T}$. In the worst case, a brute force algorithm can go through all the possible policies for agent 1 and for each one of them compute the

optimal policy for agent 2. Agent 2 builds its belief-state MDP, where each node is the agent's belief that the global state is a state $s$. There is an edge for any possible action and message that the agents can choose. Agent 2 can choose any action $a_2 \in A_2$ and it can either send a null message, or a message with its last observation ($s_2$). For any possible policy of action and communication of agent 1, agent 2 can build such a belief-state MDP and solve it. This can also be done in time that is polynomial in the number of the belief-states. Whenever an agent sends a non-null message, then the belief-state MDP has a transition to a state that is fully observable with probability one. In any case, each agent needs only to remember its last current partial view, so the complexity of solving a Dec-MDP-Com with independent observations and transitions is in the NP class. In the worst case, the cost of sending a message can be prohibitively large, therefore policies that do not send a message need also to be evaluated. It is already known (Papadimitriou & Tsitsiklis, 1982, 1986) that a simple decentralized decision-making problem for two agents is NP-hard (where $|A_i| \geq 2$ and $|A_j| \geq 3$). Therefore, the lower bound for the problem class stated in the lemma is also NP. □

The proof of this lemma is based on the number of policies that need to be evaluated by a naive algorithm. Limiting ourselves to GO-Dec-MDP with direct communication does not change the number of policies. Therefore, we obtain the same complexity for the goal-oriented case with direct communication and independent transitions and observations:

**Corollary 5** *Deciding a GO-Dec-MDP-Com with independent transitions and observations is NP-complete.*

To address the complexity of decentralized control with communication, we propose a practical approximation technique (Goldman & Zilberstein, 2004, 2003). This approach is based on meta-level control of communication, motivated by a similar decision-theoretic approach to meta-level reasoning that was developed by Russell and Wefald (1991). We assume that the designer of the system also designs a mechanism for communication. This mechanism stipulates how to decompose the global problem into local (single-agent) and temporary problems that depend on the information held by the agents whenever they exchange messages.

## 6. Discussion

Decentralized control problems are very intriguing from a theoretical point of view as well as from a practical point of view. From a theoretical perspective, decentralized partially-observable Markov decision processes serve as a formal framework to study the foundations of multi-agent systems (e.g., Becker et al., 2003; Hansen et al., 2004; Guestrin & Gordon, 2002; Peshkin et al., 2000; Pynadath & Tambe, 2002; Claus & Boutilier, 1998). Our study focuses on computing off-line decentralized policies of control for cooperative systems. This paper analyzes the complexity of solving these problems *optimally* for certain classes of decentralized control. We found critical transitions in complexity between classes of problems that range from NEXP to P. The critical transition occurs when we assume that the decentralized problems have independent transitions and observations. Adding only goal-oriented behavior to a decentralized problem does not simplify it, unless the transitions and observations are independent. We extended the decentralized process model to enable direct communication among the agents, taking into account that this communication incurs a certain cost. Communication allows the agents to synchronize their knowledge and thus eliminate the uncertainty about the global state of the world (at least at certain times). We also analyzed the complexity of the decentralized problems with indirect and direct communication.

From a practical perspective, decentralized control problems appear frequently in real-world applications where the decision-makers may be robots placed at separate geographical locations or computational processes distributed in information space. The classes of Dec-POMDPs that we identify seem to match many practical applications. Independent transitions and observations arise in examples such as multi-agent mapping, flexible manufacturing, and multiple-rovers working on data-collection in uncertain terrains, when the agents' actions are not strongly coupled. Goal-oriented behavior is relevant in these examples when the agents' behavior is aimed at reaching specific states, for example the assembly of a particular machine in a manufacturing process, or the retrieval of information.

We analyzed the notion of information sharing in decentralized systems by distinguishing among three possible sources for information: indirect communication attained by agents observing *dependent* observations, direct communication achieved by adding an external language of communication, and common uncontrollable features. The typical distinction previously made in the literature is between systems with no communication and systems with a predefined language of communication, which typically does not incur any costs, overlooking the fact that dependent observations offer yet another form of communication (Pynadath & Tambe, 2002; Decker & Lesser, 1992; Grosz & Kraus, 1996; Durfee, 1988; Roth, Vail, & Veloso, 2003). The problem of combining communication acts into the decision problem of a group of cooperative agents was addressed by Xuan et al. (2001). Their framework is similar to ours but their approach is heuristic. We proved that the language of the observations is sufficient in order to reach an optimal decentralized solution (assuming all the messages incur the same cost). This leads to the understanding that any other type of communication can serve as an approximation to the optimal solution, which may be easier to obtain.

In addition to presenting a formal framework of decentralized control, we introduced tractable algorithms for solving optimally certain classes of Dec-MDPs. The Coverage-set algorithm that solves optimally decentralized MDPs with a certain reward structure appeared in (Becker et al., 2003). Here, we add two optimal algorithms aimed at goal-oriented decentralized control.

The contribution of this paper is in framing and categorizing fundamental issues in decentralized control of cooperative systems. In particular, we characterized and studied the complexity of goal-oriented behavior, jointly fully-observable processes and independent transitions and observations, which result in important and practical classes of control problems. We also studied three sources for information sharing in such decentralized systems and provided algorithms that compute optimal solutions. Future research will focus on algorithms for decentralized control with direct communication of messages that are different from the observations. In particular, we will examine more general models of communication that allow exchange of partial information and involve unreliable communication.

## Acknowledgments

The authors wish to thank Dan Bernstein for interesting discussions on the complexity of Dec-MDPs. This work was supported in part by the National Science Foundation under grant IIS-0219606, by the Air Force Office of Scientific Research under grant F49620-03-1-0090 and by NASA under cooperative agreement NCC 2-1311. Any opinions, findings, and conclusions or recommendations expressed in this material are those of the authors and do not reflect the views of the NSF, AFOSR or NASA.